\documentclass[sigconf]{acmart}
\AtBeginDocument{%
  }

\setcopyright{acmlicensed}
\copyrightyear{2018}
\acmYear{2018}
\acmDOI{XXXXXXX.XXXXXXX}
\acmConference[Conference acronym 'XX]{Make sure to enter the correct
  conference title from your rights confirmation email}{June 03--05,
  2018}{Woodstock, NY}
\acmISBN{978-1-4503-XXXX-X/2018/06}

\begin{document}

\title{Generative Models for Synthetic Data: Transforming Data Mining in the GenAI Era}


\author{Dawei Li}
\affiliation{%
  \institution{Arizona State University}
  \city{Tempe}
  \state{Arizona}
  \country{USA}
}

\author{Yue Huang}
\affiliation{%
  \institution{University of Notre Dame}
  \city{South Bend}
  \state{Indiana}
  \country{USA}
}

\author{Ming Li}
\affiliation{%
  \institution{University of Maryland}
  \city{College Park}
  \state{Maryland}
  \country{USA}
}

\author{Tianyi Zhou}
\affiliation{%
  \institution{University of Maryland}
  \city{College Park}
  \state{Maryland}
  \country{USA}
}

\author{Xiangliang Zhang}
\affiliation{%
  \institution{University of Notre Dame}
  \city{South Bend}
  \state{Indiana}
  \country{USA}
}

\author{Huan Liu}
\affiliation{%
  \institution{Arizona State University}
  \city{Tempe}
  \state{Arizona}
  \country{USA}
}

\renewcommand{\shortauthors}{Trovato et al.}

\begin{abstract}
Generative models such as Large Language Models, Diffusion Models, and generative adversarial networks have recently revolutionized the creation of synthetic data, offering scalable solutions to data scarcity, privacy, and annotation challenges in data mining. This tutorial introduces the foundations and latest advances in synthetic data generation, covers key methodologies and practical frameworks, and discusses evaluation strategies and applications. Attendees will gain actionable insights into leveraging generative synthetic data to enhance data mining research and practice. More information can be found on our website: \url{https://syndata4dm.github.io/}.
\end{abstract}




\received{20 February 2007}
\received[revised]{12 March 2009}
\received[accepted]{5 June 2009}

\maketitle

\section{INTRODUCTION}

In the era of data-driven artificial intelligence (AI), access to large-scale, high-quality datasets has become a fundamental requirement for breakthroughs in data mining and machine learning. However, real-world data is often scarce, expensive to annotate, or restricted due to privacy and proprietary concerns. Synthetic data, algorithmically generated datasets that mimic the statistical properties and underlying patterns of real-world data, has emerged as a powerful solution to these challenges.

\begin{figure}
    \centering
    \includegraphics[width=1.0\linewidth]{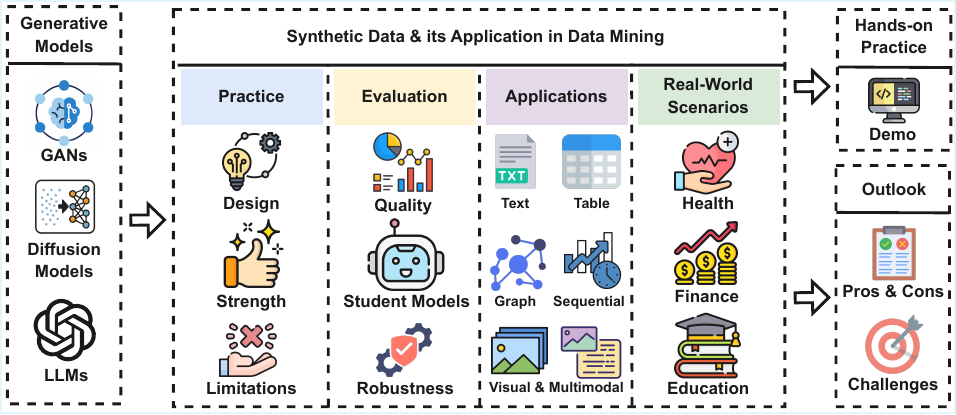}
    \caption{The overview of our tutorial.}
    \label{fig:intro}
    \vspace{-15pt}
\end{figure}

Recent advances in generative models, such as Large Language Models (LLMs), Diffusion Models, and generative adversarial networks (GANs) have significantly enhanced our ability to generate realistic, diverse, and controllable synthetic data across a wide range of data types. Synthetic data powered by these generative models is revolutionizing the way we approach data mining, from augmenting training datasets and reducing annotation costs to enabling privacy-preserving analytics and fostering innovation in low-resource or long-tail scenarios.

This tutorial aims to provide an up-to-date overview of the foundations, methodologies, practical frameworks, evaluation techniques, and real-world applications of synthetic data generation using modern generative models. We will highlight both the opportunities and challenges in leveraging synthetic data for data mining, empowering researchers and practitioners to effectively apply these cutting-edge techniques in their own domains.

\section{TARGET AUDIENCE, PREREQUISITES, AND BENEFITS}

\paragraph{Target Audience.} This tutorial is intended for researchers, practitioners, and graduate students in data mining, machine learning, natural language processing, and computer vision who are interested in using synthetic data to address challenges such as data scarcity, privacy, and generalization.

\paragraph{Prerequisites.} Attendees are expected to have a basic understanding of machine learning and deep learning, and some familiarity with generative models such as Large Language Models and Diffusion Models. Prior exposure to NLP, computer vision, or multimodal tasks is helpful but not required.

\paragraph{Benefits.} By the end of the tutorial, participants will have a clear overview of recent advances in synthetic data generation, learn how to apply existing tools to real-world problems, and understand the practical considerations involved in evaluating and using synthetic data effectively.

\section{OUTLINE of TUTORIAL}


Our tutorial is designed as a half-day session, spanning 3 hours with a short break in between. The detailed agenda is as follows:
\paragraph{Part 1} Foundations of Synthetic Data (85 mins)
    \begin{itemize}
        \item Introduction \& Background (15 mins)
        \item Core Generative Models for Data Generation (30 mins)
        \item Synthetic Data in Practice (20 mins)
        \item Evaluation and Benchmarking (20 mins)
    \end{itemize}
\paragraph{Coffee Break} (15 mins)
\paragraph{Part 2} Applications and Future Directions (80 mins)
    \begin{itemize}
        \item Applications in Data Mining (30 mins)
        \item Real-world Scenarios (15 mins)
        \item Hands-on Practice (20 mins)
        \item Challenges, Future Directions, and Q\&A (15 mins)
    \end{itemize}

\subsection{Introduction \& Background}

\paragraph{Definition \& Motivation}

In this section, we will first introduce the definition and motivation of synthetic data generation. In the broadest sense, it refers to the process of generating artificial data that algorithmically emulates the statistical properties and underlying patterns of a real-world dataset. In today’s AI landscape, the cost and constraints of curating large-scale, high-quality datasets have become a primary bottleneck to progress. Thus, we will include the motivation of synthetic data generation from different aspects \cite{tan2024large, nadas2025synthetic}, including: data scarcity, annotation cost reduction, privacy preservation, proprietary data protection, long-tail distributions, low-resource scenarios, etc. 

\paragraph{Background}
We will also include the background knowledge regarding synthetic data generation, from early rule-based augmentation or GAN-style generators that worked well in vision to advanced techniques utilizing Diffusion Models and Large Language Models. Also, we will provide an overview of the real-world applications of synthetic data generation and how it relates to the area of data mining. In this section, we hope to provide a quick glance at the topic to the audience. 

\subsection{Core Generative Models for Synthetic Data Generation}

We will mainly cover 3 categories as the core generative models: Generative Adversarial Networks (GANs), Diffusion Models, and Large Language Models. 

\paragraph{Generative Adversarial Networks}
GANs \cite{goodfellow2014generative} remain the ``classics'' of synthetic data generation: a generator learns to fool a discriminator that distinguishes real from fake, gradually shaping samples that follow the true data manifold. We will introduce how vanilla GANs work, some milestone GAN-style models, including Style-GAN \cite{karras2019style}, Drag-GAN \cite{pan2023drag}, etc. We will introduce their pros and cons, and why they are not as popular as the 2 categories below. 

\paragraph{Diffusion Models}
Diffusion \cite{ho2020denoising} treats generation as incremental denoising: data are gradually noised to a latent prior, then reconstructed step-by-step with a learnable reverse process. We will introduce how vanilla Diffusion Models work, some milestones, including DALL·E, Stable Diffusion \cite{rombach2022high}, Sora, etc. We will compare their pros and cons with GAN-style models.  

\paragraph{Large Language Models}
Instruction-tuned LLMs have revolutionized text-centric synthesis: a single prompt can emit grammatically correct paragraphs. Different from the above 2 categories, this series of models focuses mainly on text data generation, either for texts as queries or images, videos as queries. We will introduce milestone models and methods for this section.

\subsection{Synthetic Data in Practice}

In this section, we will introduce some of the most recent and advanced frameworks for synthetic data generation. For text-based data, we will discuss systems such as MagPie \cite{xu2024magpie}, DataGen \cite{huang2024datagen}, and DyVal \cite{zhu2023dyval, zhu2024dyval}. For multimodal data (e.g., text-image or text-audio pairs), we will cover frameworks like Task-Me-Anything \cite{zhangtask} and AutoBench-v \cite{bao2024autobench}.

We will explore how these frameworks are designed, the underlying generative techniques they employ, and how they address different challenges in data synthesis. Furthermore, we will analyze their respective strengths, such as scalability, controllability, and data diversity, as well as their limitations, including potential biases, domain generalization issues, and computational overhead.

\subsection{Evaluation and Benchmarking}

Evaluating synthetic data is a critical yet challenging task in machine learning and data science. Current approaches assess data fidelity, diversity, controllability, truthfulness, and downstream utility through a combination of quantitative metrics and task-specific performance on real-world benchmarks \cite{maheshwari2024efficacy, huang2024datagen, kim2024evaluating}. In practice, synthetic data is often evaluated by training models on generated datasets and then measuring their performance on downstream tasks, which serves as a proxy for real-world applicability. Despite these advances, robust and interpretable evaluation remains an open problem. In particular, existing methods struggle to comprehensively address issues such as data bias \cite{ye2024justice}, ethical risks, and the generalization capabilities of synthetic data across different domains and applications.

\subsection{Applications in Data Mining}
\label{Applications in Data Mining}

\paragraph{Text Data.}
Synthetic text data enhances text mining tasks such as classification, relation extraction, and named entity recognition. Approaches mainly include: (1) generating or augmenting input text to enrich datasets~\cite{yu2023large}, and (2) generating pseudo labels for unlabeled data to facilitate annotation~\cite{zhang2023llmaaa}, improving both data diversity and efficiency.

\paragraph{Tabular Data.}
Tabular data synthesis involves: (1) generative modeling with diffusion, flow-based, or GAN-based models~\cite{kotelnikov2023tabddpm}, (2) conditional table generation guided by schema and control signals~\cite{shi2025tabdiff,lin2024ctsyn}, and (3) table extraction from raw text using LLMs~\cite{liu2024discoveryhiddenworldlarge}. Synthetic tables support privacy-preserving release, data augmentation, and robust learning.

\paragraph{Graph Data.}
Graph data synthesis advances molecule, protein, network analysis, and knowledge graph construction~\cite{liugenerative,li2024dalk}. Key approaches include: (1) structure-level generation of graph topologies~\cite{jo2023graph}, (2) node/edge-level augmentation~\cite{wang2025diffusion}, and (3) conditional generation from textual or structured input~\cite{yaotext}.

\paragraph{Sequential Data.}
Sequential data synthesis includes: (1) time series generation that captures complex temporal patterns~\cite{liu2025empowering}, and (2) representation synthesis for sequential recommendation, augmenting user-item interactions~\cite{li2023diffurec}. These techniques balance class distributions, simulate rare events, and aid pretraining.

\paragraph{Visual \& Multimodal Data.}
Visual and multimodal synthesis spans: (1) image generation with foundation diffusion models (e.g., Stable Diffusion, DALL-E), and (2) multimodal generation of aligned visual-language pairs. Synthetic data enables efficient creation of diverse, labeled datasets for improved training of vision and multimodal models.

\subsection{Real-world Scenarios}
In this section, we will discuss the utilization of synthetic data in various real-world data mining scenarios, including health, finance, and education. In the health domain, foundation models such as GPT-4 and hierarchical autoregressive transformers have been employed to generate synthetic clinical records and electronic health data, supporting tasks like named entity recognition and patient outcome prediction while preserving privacy~\cite{theodorou2023synthesize}. In finance, generative models including Diffusion Models and GANs have been used to simulate realistic transaction data for fraud detection, address data imbalance, and facilitate data sharing among financial institutions under privacy constraints~\cite{pushkarenko2024synthetic}. In the education sector, synthetic student performance records produced by GANs and LLMs have enabled accurate predictive modeling for student outcomes while mitigating data scarcity issues~\cite{farhood2024advancing}.

\subsection{Hands-on Practice}
In this section, we aim to provide a demon program as hands-on practice to show how to synthesize each of the types of synthetic data we mentioned in Section~\ref{Applications in Data Mining}, including text~\cite{yu2023large}, tabular~\cite{shi2025tabdiff}, graph~\cite{li2024dalk}, sequential~\cite{liu2025empowering} and visual \& multimodal~\cite{tian2023stablerep} data.

\subsection{Outlook}
\paragraph{Pros \& Cons}
In this section, we outline both the pros and cons of the use of synthetic data in data mining application. The pros include: (1). enhancing data privacy by avoiding real personal records, (2). enabling large-scale data generation and (3). helping address data imbalances~\cite{ba2024fill}. The cons involves: (1). failing to capture the nuances of real-world distributions, (2). learn spurious or unrealistic patterns, and (3). overfitting to the artificial distribution~\cite{hao2024synthetic}.

\paragraph{Challenges and Future Directions.}
In this section, we highlight several challenges and promising directions for future research in this area. Although model collapse~\cite{shumailov2024ai} has been observed in generative models trained iteratively on synthetic data, its effects on the data distribution of data mining models remain underexplored and warrant further investigation. Moreover, there is still a lack of effective strategies for integrating generative model-based and traditional data synthesis methods. Such integration could enable the generation of more trustworthy synthetic data across scenarios.

\section{RELATED TUTORIALS}

\href{https://2025.aclweb.org/program/tutorials/}{Synthetic Data in the Era of Large Language Models @ ACL 2025}, 

\noindent\href{https://sites.google.com/view/synthetic-tab-data-tutorial}{Synthetic Tabular Data: methods, attacks and defenses @ KDD 2025}

\noindent\href{https://icml.cc/virtual/2021/tutorial/10846}{Synthetic Healthcare Data Generation and Assessment: Challenges, Methods, and Impact on Machine Learning @ ICML 2021}

While these tutorials each focus on a single modality or application, our session uniquely spans all major data types—text, tabular, graph, sequential, and multimodal—showcasing the latest generative models, end-to-end frameworks, and unified evaluation strategies tailored for diverse data mining tasks.

\section{PRESENTER BIOGRAPHY}

\textbf{Dawei Li} is a Ph.D. student in Computer Science at Arizona State University. Previously, He obtained his bachelor’s degree in Computer Science from Beijing Language and Cultural University and master’s degree in Data Science from the University of California, San Diego. His research focuses on techniques and risks from AI oversight. Dawei have published papers and served as reviewers in top NLP and Data Mining venues including ACL, EMNLP, NAACL, TKDD, PAKDD and SIGKDD Exploration.

\textbf{Yue Huang} is a Ph.D. student in Computer Science and Engineering at the University of Notre Dame. He earned his B.S.\ in Computer Science from Sichuan University. His research investigates the trustworthiness and social responsibility of foundation models. Yue has published extensively at premier venues including NeurIPS, ICLR, ICML, ACL, EMNLP, NAACL, CVPR, and IJCAI. His work has been highlighted by the U.S.\ Department of Homeland Security and recognized with the Microsoft Accelerating Foundation Models Research Award and the KAUST AI Rising Star Award (2025). 
\looseness-1

\textbf{Ming Li} is a Ph.D. student in Computer Science at the University of Maryland. Previously, He obtained his bachelor’s degree in Computer Science from Xi'an Jiaotong University and his master’s degree in Computer Science from Texas A\&M University. 
His research focuses on post-training for foundation models and responsible and self-evolving AI. 
Ming has published papers and served as a reviewer in top NLP and Machine Learning venues, including ACL, EMNLP, ICLR, NAACL, and etc.

\textbf{Tianyi Zhou} is a tenure-track assistant professor of Computer Science at the University of Maryland, College Park (UMD). He received his Ph.D. from the University of Washington and worked as a research scientist at Google before joining UMD. His research interests are machine learning, natural language processing, and multi-modal generative AI. His team has published >130 papers in ML (NeurIPS, ICML, ICLR), NLP (ACL, EMNLP, NAACL), CV (CVPR, ICCV, ECCV), and journals such as IEEE TPAMI/TIP/TNNLS/TKDE, with >10000 citations. He is the recipient of the best student paper of ICDM 2013. He has been serving as an area chair of ICLR, NeurIPS, ACL, EMNLP, SIGKDD, AAAI, IJCAI, WACV, etc.

\textbf{Xiangliang Zhang} is a Leonard C. Bettex Collegiate Professor in the Department of Computer Science and Engineering, University of Notre Dame. She was an Associate Professor in Computer Science at the King Abdullah University of Science and Technology (KAUST), Saudi Arabia. She received her Ph.D. degree in computer science from INRIA-Universite Paris Sud, France, in 2010. Her main research interests and experiences are in machine learning and data mining. She has published more than 270 refereed papers in leading international conferences and journals. She serves as associate editor of IEEE Transactions on Dependable and Secure Computing, Information Sciences, and International Journal of Intelligent Systems, and regularly serves as area chair or on the (senior) program committee of IJCAI, SIGKDD, NeurIPS, AAAI, ICML, and WSDM.

\textbf{Huan Liu} is a Regent Professor in the School of Computing, and Augmented Intelligence, Arizona State University. He received his Ph.D. degree in Computer Science from the University of Southern California, in 1989. His research focuses on developing computational methods for data mining, machine learning, and social computing. Dr. Liu has been honored with numerous prestigious awards: ACM SIGKDD Innovation Award (2022) for his pioneering work in feature selection and social media mining, Fellow of ACM (2018), AAAI (2019), AAAS (2018), and IEEE (2012). He is Chief Editor of ACM TIST, Frontiers in Big Data and DMM, and has been actively involved on editorial boards and program committees for major conferences such as KDD, ICML, NeurIPS, AAAI, and IJCAI.
\bibliographystyle{ACM-Reference-Format}
\bibliography{sample-base, software}


\begin{thebibliography}{36}


\ifx \showCODEN    \undefined \def \showCODEN     #1{\unskip}     \fi
\ifx \showISBNx    \undefined \def \showISBNx     #1{\unskip}     \fi
\ifx \showISBNxiii \undefined \def \showISBNxiii  #1{\unskip}     \fi
\ifx \showISSN     \undefined \def \showISSN      #1{\unskip}     \fi
\ifx \showLCCN     \undefined \def \showLCCN      #1{\unskip}     \fi
\ifx \shownote     \undefined \def \shownote      #1{#1}          \fi
\ifx \showarticletitle \undefined \def \showarticletitle #1{#1}   \fi
\ifx \showURL      \undefined \def \showURL       {\relax}        \fi
\providecommand\bibfield[2]{#2}
\providecommand\bibinfo[2]{#2}
\providecommand\natexlab[1]{#1}
\providecommand\showeprint[2][]{arXiv:#2}

\bibitem[Ba et~al\mbox{.}(2024)]%
        {ba2024fill}
\bibfield{author}{\bibinfo{person}{Yang Ba}, \bibinfo{person}{Michelle~V Mancenido}, {and} \bibinfo{person}{Rong Pan}.} \bibinfo{year}{2024}\natexlab{}.
\newblock \showarticletitle{Fill In The Gaps: Model Calibration and Generalization with Synthetic Data}.
\newblock \bibinfo{journal}{\emph{arXiv preprint arXiv:2410.10864}} (\bibinfo{year}{2024}).
\newblock


\bibitem[Bao et~al\mbox{.}(2024)]%
        {bao2024autobench}
\bibfield{author}{\bibinfo{person}{Han Bao}, \bibinfo{person}{Yue Huang}, \bibinfo{person}{Yanbo Wang}, \bibinfo{person}{Jiayi Ye}, \bibinfo{person}{Xiangqi Wang}, \bibinfo{person}{Xiuying Chen}, \bibinfo{person}{Yue Zhao}, \bibinfo{person}{Tianyi Zhou}, \bibinfo{person}{Mohamed Elhoseiny}, {and} \bibinfo{person}{Xiangliang Zhang}.} \bibinfo{year}{2024}\natexlab{}.
\newblock \showarticletitle{AutoBench-V: Can Large Vision-Language Models Benchmark Themselves?}
\newblock \bibinfo{journal}{\emph{arXiv preprint arXiv:2410.21259}} (\bibinfo{year}{2024}).
\newblock


\bibitem[Farhood et~al\mbox{.}(2024)]%
        {farhood2024advancing}
\bibfield{author}{\bibinfo{person}{Helia Farhood}, \bibinfo{person}{Ibrahim Joudah}, \bibinfo{person}{Amin Beheshti}, {and} \bibinfo{person}{Samuel Muller}.} \bibinfo{year}{2024}\natexlab{}.
\newblock \showarticletitle{Advancing student outcome predictions through generative adversarial networks}.
\newblock \bibinfo{journal}{\emph{Computers and Education: Artificial Intelligence}}  \bibinfo{volume}{7} (\bibinfo{year}{2024}), \bibinfo{pages}{100293}.
\newblock


\bibitem[Goodfellow et~al\mbox{.}(2014)]%
        {goodfellow2014generative}
\bibfield{author}{\bibinfo{person}{Ian~J Goodfellow}, \bibinfo{person}{Jean Pouget-Abadie}, \bibinfo{person}{Mehdi Mirza}, \bibinfo{person}{Bing Xu}, \bibinfo{person}{David Warde-Farley}, \bibinfo{person}{Sherjil Ozair}, \bibinfo{person}{Aaron Courville}, {and} \bibinfo{person}{Yoshua Bengio}.} \bibinfo{year}{2014}\natexlab{}.
\newblock \showarticletitle{Generative adversarial nets}.
\newblock \bibinfo{journal}{\emph{Advances in neural information processing systems}}  \bibinfo{volume}{27} (\bibinfo{year}{2014}).
\newblock


\bibitem[Hao et~al\mbox{.}(2024)]%
        {hao2024synthetic}
\bibfield{author}{\bibinfo{person}{Shuang Hao}, \bibinfo{person}{Wenfeng Han}, \bibinfo{person}{Tao Jiang}, \bibinfo{person}{Yiping Li}, \bibinfo{person}{Haonan Wu}, \bibinfo{person}{Chunlin Zhong}, \bibinfo{person}{Zhangjun Zhou}, {and} \bibinfo{person}{He Tang}.} \bibinfo{year}{2024}\natexlab{}.
\newblock \showarticletitle{Synthetic data in AI: Challenges, applications, and ethical implications}.
\newblock \bibinfo{journal}{\emph{arXiv preprint arXiv:2401.01629}} (\bibinfo{year}{2024}).
\newblock


\bibitem[Ho et~al\mbox{.}(2020)]%
        {ho2020denoising}
\bibfield{author}{\bibinfo{person}{Jonathan Ho}, \bibinfo{person}{Ajay Jain}, {and} \bibinfo{person}{Pieter Abbeel}.} \bibinfo{year}{2020}\natexlab{}.
\newblock \showarticletitle{Denoising diffusion probabilistic models}.
\newblock \bibinfo{journal}{\emph{Advances in neural information processing systems}}  \bibinfo{volume}{33} (\bibinfo{year}{2020}), \bibinfo{pages}{6840--6851}.
\newblock


\bibitem[Huang et~al\mbox{.}(2024)]%
        {huang2024datagen}
\bibfield{author}{\bibinfo{person}{Yue Huang}, \bibinfo{person}{Siyuan Wu}, \bibinfo{person}{Chujie Gao}, \bibinfo{person}{Dongping Chen}, \bibinfo{person}{Qihui Zhang}, \bibinfo{person}{Yao Wan}, \bibinfo{person}{Tianyi Zhou}, \bibinfo{person}{Chaowei Xiao}, \bibinfo{person}{Jianfeng Gao}, \bibinfo{person}{Lichao Sun}, {et~al\mbox{.}}} \bibinfo{year}{2024}\natexlab{}.
\newblock \showarticletitle{Datagen: Unified synthetic dataset generation via large language models}. In \bibinfo{booktitle}{\emph{The Thirteenth International Conference on Learning Representations}}.
\newblock


\bibitem[Jo et~al\mbox{.}(2023)]%
        {jo2023graph}
\bibfield{author}{\bibinfo{person}{Jaehyeong Jo}, \bibinfo{person}{Dongki Kim}, {and} \bibinfo{person}{Sung~Ju Hwang}.} \bibinfo{year}{2023}\natexlab{}.
\newblock \showarticletitle{Graph generation with diffusion mixture}.
\newblock \bibinfo{journal}{\emph{arXiv preprint arXiv:2302.03596}} (\bibinfo{year}{2023}).
\newblock


\bibitem[Karras et~al\mbox{.}(2019)]%
        {karras2019style}
\bibfield{author}{\bibinfo{person}{Tero Karras}, \bibinfo{person}{Samuli Laine}, {and} \bibinfo{person}{Timo Aila}.} \bibinfo{year}{2019}\natexlab{}.
\newblock \showarticletitle{A style-based generator architecture for generative adversarial networks}. In \bibinfo{booktitle}{\emph{Proceedings of the IEEE/CVF conference on computer vision and pattern recognition}}. \bibinfo{pages}{4401--4410}.
\newblock


\bibitem[Kim et~al\mbox{.}(2024)]%
        {kim2024evaluating}
\bibfield{author}{\bibinfo{person}{Seungone Kim}, \bibinfo{person}{Juyoung Suk}, \bibinfo{person}{Xiang Yue}, \bibinfo{person}{Vijay Viswanathan}, \bibinfo{person}{Seongyun Lee}, \bibinfo{person}{Yizhong Wang}, \bibinfo{person}{Kiril Gashteovski}, \bibinfo{person}{Carolin Lawrence}, \bibinfo{person}{Sean Welleck}, {and} \bibinfo{person}{Graham Neubig}.} \bibinfo{year}{2024}\natexlab{}.
\newblock \showarticletitle{Evaluating Language Models as Synthetic Data Generators}.
\newblock \bibinfo{journal}{\emph{CoRR}}  \bibinfo{volume}{abs/2412.03679} (\bibinfo{year}{2024}).
\newblock
\showeprint[arxiv]{2412.03679}
\newblock
\shownote{preprint}.


\bibitem[Kotelnikov et~al\mbox{.}(2023)]%
        {kotelnikov2023tabddpm}
\bibfield{author}{\bibinfo{person}{Akim Kotelnikov}, \bibinfo{person}{Dmitry Baranchuk}, \bibinfo{person}{Ivan Rubachev}, {and} \bibinfo{person}{Artem Babenko}.} \bibinfo{year}{2023}\natexlab{}.
\newblock \showarticletitle{Tabddpm: Modelling tabular data with diffusion models}. In \bibinfo{booktitle}{\emph{International Conference on Machine Learning}}. PMLR, \bibinfo{pages}{17564--17579}.
\newblock


\bibitem[Li et~al\mbox{.}(2024)]%
        {li2024dalk}
\bibfield{author}{\bibinfo{person}{Dawei Li}, \bibinfo{person}{Shu Yang}, \bibinfo{person}{Zhen Tan}, \bibinfo{person}{Jae Baik}, \bibinfo{person}{Sukwon Yun}, \bibinfo{person}{Joseph Lee}, \bibinfo{person}{Aaron Chacko}, \bibinfo{person}{Bojian Hou}, \bibinfo{person}{Duy Duong-Tran}, \bibinfo{person}{Ying Ding}, {et~al\mbox{.}}} \bibinfo{year}{2024}\natexlab{}.
\newblock \showarticletitle{DALK: Dynamic Co-Augmentation of LLMs and KG to answer Alzheimer’s Disease Questions with Scientific Literature}. In \bibinfo{booktitle}{\emph{Findings of the Association for Computational Linguistics: EMNLP 2024}}. \bibinfo{pages}{2187--2205}.
\newblock


\bibitem[Li et~al\mbox{.}(2023)]%
        {li2023diffurec}
\bibfield{author}{\bibinfo{person}{Zihao Li}, \bibinfo{person}{Aixin Sun}, {and} \bibinfo{person}{Chenliang Li}.} \bibinfo{year}{2023}\natexlab{}.
\newblock \showarticletitle{Diffurec: A diffusion model for sequential recommendation}.
\newblock \bibinfo{journal}{\emph{ACM Transactions on Information Systems}} \bibinfo{volume}{42}, \bibinfo{number}{3} (\bibinfo{year}{2023}), \bibinfo{pages}{1--28}.
\newblock


\bibitem[Lin et~al\mbox{.}(2024)]%
        {lin2024ctsyn}
\bibfield{author}{\bibinfo{person}{Xiaofeng Lin}, \bibinfo{person}{Chenheng Xu}, \bibinfo{person}{Matthew Yang}, {and} \bibinfo{person}{Guang Cheng}.} \bibinfo{year}{2024}\natexlab{}.
\newblock \showarticletitle{CTSyn: A Foundational Model for Cross Tabular Data Generation}.
\newblock \bibinfo{journal}{\emph{arXiv preprint arXiv:2406.04619}} (\bibinfo{year}{2024}).
\newblock


\bibitem[Liu et~al\mbox{.}(2024)]%
        {liu2024discoveryhiddenworldlarge}
\bibfield{author}{\bibinfo{person}{Chenxi Liu}, \bibinfo{person}{Yongqiang Chen}, \bibinfo{person}{Tongliang Liu}, \bibinfo{person}{Mingming Gong}, \bibinfo{person}{James Cheng}, \bibinfo{person}{Bo Han}, {and} \bibinfo{person}{Kun Zhang}.} \bibinfo{year}{2024}\natexlab{}.
\newblock \showarticletitle{Discovery of the Hidden World with Large Language Models}.
\newblock  (\bibinfo{year}{2024}).
\newblock
\showeprint[arxiv]{2402.03941}~[cs.LG]
\urldef\tempurl%
\url{https://arxiv.org/abs/2402.03941}
\showURL{%
\tempurl}


\bibitem[Liu et~al\mbox{.}({[n.\,d.]})]%
        {liugenerative}
\bibfield{author}{\bibinfo{person}{Chengyi Liu}, \bibinfo{person}{Wenqi Fan}, \bibinfo{person}{Yunqing Liu}, \bibinfo{person}{Jiatong Li}, \bibinfo{person}{Hang Li}, \bibinfo{person}{Hui Liu}, \bibinfo{person}{Jiliang Tang}, {and} \bibinfo{person}{Qing Li}.} \bibinfo{year}{[n.\,d.]}\natexlab{}.
\newblock \showarticletitle{Generative Diffusion Models on Graphs: Methods and Applications}.
\newblock  (\bibinfo{year}{[n.\,d.]}).
\newblock


\bibitem[Liu et~al\mbox{.}(2025)]%
        {liu2025empowering}
\bibfield{author}{\bibinfo{person}{Xu Liu}, \bibinfo{person}{Taha Aksu}, \bibinfo{person}{Juncheng Liu}, \bibinfo{person}{Qingsong Wen}, \bibinfo{person}{Yuxuan Liang}, \bibinfo{person}{Caiming Xiong}, \bibinfo{person}{Silvio Savarese}, \bibinfo{person}{Doyen Sahoo}, \bibinfo{person}{Junnan Li}, {and} \bibinfo{person}{Chenghao Liu}.} \bibinfo{year}{2025}\natexlab{}.
\newblock \showarticletitle{Empowering Time Series Analysis with Synthetic Data: A Survey and Outlook in the Era of Foundation Models}.
\newblock \bibinfo{journal}{\emph{arXiv preprint arXiv:2503.11411}} (\bibinfo{year}{2025}).
\newblock


\bibitem[Maheshwari et~al\mbox{.}(2024)]%
        {maheshwari2024efficacy}
\bibfield{author}{\bibinfo{person}{Gaurav Maheshwari}, \bibinfo{person}{Dmitry Ivanov}, {and} \bibinfo{person}{Kevin El~Haddad}.} \bibinfo{year}{2024}\natexlab{}.
\newblock \showarticletitle{Efficacy of Synthetic Data as a Benchmark}.
\newblock \bibinfo{journal}{\emph{CoRR}}  \bibinfo{volume}{abs/2409.11968} (\bibinfo{year}{2024}).
\newblock
\showeprint[arxiv]{2409.11968}
\newblock
\shownote{preprint}.


\bibitem[Nadas et~al\mbox{.}(2025)]%
        {nadas2025synthetic}
\bibfield{author}{\bibinfo{person}{Mihai Nadas}, \bibinfo{person}{Laura Diosan}, {and} \bibinfo{person}{Andreea Tomescu}.} \bibinfo{year}{2025}\natexlab{}.
\newblock \showarticletitle{Synthetic data generation using large language models: Advances in text and code}.
\newblock \bibinfo{journal}{\emph{arXiv preprint arXiv:2503.14023}} (\bibinfo{year}{2025}).
\newblock


\bibitem[Pan et~al\mbox{.}(2023)]%
        {pan2023drag}
\bibfield{author}{\bibinfo{person}{Xingang Pan}, \bibinfo{person}{Ayush Tewari}, \bibinfo{person}{Thomas Leimk{\"u}hler}, \bibinfo{person}{Lingjie Liu}, \bibinfo{person}{Abhimitra Meka}, {and} \bibinfo{person}{Christian Theobalt}.} \bibinfo{year}{2023}\natexlab{}.
\newblock \showarticletitle{Drag your gan: Interactive point-based manipulation on the generative image manifold}. In \bibinfo{booktitle}{\emph{ACM SIGGRAPH 2023 conference proceedings}}. \bibinfo{pages}{1--11}.
\newblock


\bibitem[Pushkarenko and Zaslavskyi(2024)]%
        {pushkarenko2024synthetic}
\bibfield{author}{\bibinfo{person}{Yurii Pushkarenko} {and} \bibinfo{person}{Volodymyr Zaslavskyi}.} \bibinfo{year}{2024}\natexlab{}.
\newblock \showarticletitle{Synthetic Data Generation for Fraud Detection Using Diffusion Models}.
\newblock \bibinfo{journal}{\emph{Information \& Security}} \bibinfo{volume}{55}, \bibinfo{number}{2} (\bibinfo{year}{2024}), \bibinfo{pages}{185--198}.
\newblock


\bibitem[Rombach et~al\mbox{.}(2022)]%
        {rombach2022high}
\bibfield{author}{\bibinfo{person}{Robin Rombach}, \bibinfo{person}{Andreas Blattmann}, \bibinfo{person}{Dominik Lorenz}, \bibinfo{person}{Patrick Esser}, {and} \bibinfo{person}{Bj{\"o}rn Ommer}.} \bibinfo{year}{2022}\natexlab{}.
\newblock \showarticletitle{High-resolution image synthesis with latent diffusion models}. In \bibinfo{booktitle}{\emph{Proceedings of the IEEE/CVF conference on computer vision and pattern recognition}}. \bibinfo{pages}{10684--10695}.
\newblock


\bibitem[Shi et~al\mbox{.}(2025)]%
        {shi2025tabdiff}
\bibfield{author}{\bibinfo{person}{Juntong Shi}, \bibinfo{person}{Minkai Xu}, \bibinfo{person}{Harper Hua}, \bibinfo{person}{Hengrui Zhang}, \bibinfo{person}{Stefano Ermon}, {and} \bibinfo{person}{Jure Leskovec}.} \bibinfo{year}{2025}\natexlab{}.
\newblock \showarticletitle{TabDiff: a Mixed-type Diffusion Model for Tabular Data Generation}. In \bibinfo{booktitle}{\emph{The Thirteenth International Conference on Learning Representations}}.
\newblock


\bibitem[Shumailov et~al\mbox{.}(2024)]%
        {shumailov2024ai}
\bibfield{author}{\bibinfo{person}{Ilia Shumailov}, \bibinfo{person}{Zakhar Shumaylov}, \bibinfo{person}{Yiren Zhao}, \bibinfo{person}{Nicolas Papernot}, \bibinfo{person}{Ross Anderson}, {and} \bibinfo{person}{Yarin Gal}.} \bibinfo{year}{2024}\natexlab{}.
\newblock \showarticletitle{AI models collapse when trained on recursively generated data}.
\newblock \bibinfo{journal}{\emph{Nature}} \bibinfo{volume}{631}, \bibinfo{number}{8022} (\bibinfo{year}{2024}), \bibinfo{pages}{755--759}.
\newblock


\bibitem[Tan et~al\mbox{.}(2024)]%
        {tan2024large}
\bibfield{author}{\bibinfo{person}{Zhen Tan}, \bibinfo{person}{Dawei Li}, \bibinfo{person}{Song Wang}, \bibinfo{person}{Alimohammad Beigi}, \bibinfo{person}{Bohan Jiang}, \bibinfo{person}{Amrita Bhattacharjee}, \bibinfo{person}{Mansooreh Karami}, \bibinfo{person}{Jundong Li}, \bibinfo{person}{Lu Cheng}, {and} \bibinfo{person}{Huan Liu}.} \bibinfo{year}{2024}\natexlab{}.
\newblock \showarticletitle{Large language models for data annotation and synthesis: A survey}.
\newblock \bibinfo{journal}{\emph{arXiv preprint arXiv:2402.13446}} (\bibinfo{year}{2024}).
\newblock


\bibitem[Theodorou et~al\mbox{.}(2023)]%
        {theodorou2023synthesize}
\bibfield{author}{\bibinfo{person}{Brandon Theodorou}, \bibinfo{person}{Cao Xiao}, {and} \bibinfo{person}{Jimeng Sun}.} \bibinfo{year}{2023}\natexlab{}.
\newblock \showarticletitle{Synthesize high-dimensional longitudinal electronic health records via hierarchical autoregressive language model}.
\newblock \bibinfo{journal}{\emph{Nature communications}} \bibinfo{volume}{14}, \bibinfo{number}{1} (\bibinfo{year}{2023}), \bibinfo{pages}{5305}.
\newblock


\bibitem[Tian et~al\mbox{.}(2023)]%
        {tian2023stablerep}
\bibfield{author}{\bibinfo{person}{Yonglong Tian}, \bibinfo{person}{Lijie Fan}, \bibinfo{person}{Phillip Isola}, \bibinfo{person}{Huiwen Chang}, {and} \bibinfo{person}{Dilip Krishnan}.} \bibinfo{year}{2023}\natexlab{}.
\newblock \showarticletitle{Stablerep: Synthetic images from text-to-image models make strong visual representation learners}.
\newblock \bibinfo{journal}{\emph{Advances in Neural Information Processing Systems}}  \bibinfo{volume}{36} (\bibinfo{year}{2023}), \bibinfo{pages}{48382--48402}.
\newblock


\bibitem[Wang et~al\mbox{.}(2025)]%
        {wang2025diffusion}
\bibfield{author}{\bibinfo{person}{Yancheng Wang}, \bibinfo{person}{Changyu Liu}, {and} \bibinfo{person}{Yingzhen Yang}.} \bibinfo{year}{2025}\natexlab{}.
\newblock \showarticletitle{Diffusion on Graph: Augmentation of Graph Structure for Node Classification}.
\newblock \bibinfo{journal}{\emph{arXiv preprint arXiv:2503.12563}} (\bibinfo{year}{2025}).
\newblock


\bibitem[Xu et~al\mbox{.}(2024)]%
        {xu2024magpie}
\bibfield{author}{\bibinfo{person}{Zhangchen Xu}, \bibinfo{person}{Fengqing Jiang}, \bibinfo{person}{Luyao Niu}, \bibinfo{person}{Yuntian Deng}, \bibinfo{person}{Radha Poovendran}, \bibinfo{person}{Yejin Choi}, {and} \bibinfo{person}{Bill~Yuchen Lin}.} \bibinfo{year}{2024}\natexlab{}.
\newblock \showarticletitle{Magpie: Alignment data synthesis from scratch by prompting aligned llms with nothing}.
\newblock \bibinfo{journal}{\emph{arXiv preprint arXiv:2406.08464}} (\bibinfo{year}{2024}).
\newblock


\bibitem[Yao et~al\mbox{.}({[n.\,d.]})]%
        {yaotext}
\bibfield{author}{\bibinfo{person}{Yang Yao}, \bibinfo{person}{Xin Wang}, \bibinfo{person}{Yijian Qin}, \bibinfo{person}{Zeyang Zhang}, \bibinfo{person}{Wenwu Zhu}, {and} \bibinfo{person}{Hong Mei}.} \bibinfo{year}{[n.\,d.]}\natexlab{}.
\newblock \showarticletitle{Text-to-graph Generation with Conditional Diffusion Models Guided by Graph-aligned LLMs}.
\newblock  (\bibinfo{year}{[n.\,d.]}).
\newblock


\bibitem[Ye et~al\mbox{.}(2024)]%
        {ye2024justice}
\bibfield{author}{\bibinfo{person}{Jiayi Ye}, \bibinfo{person}{Yanbo Wang}, \bibinfo{person}{Yue Huang}, \bibinfo{person}{Dongping Chen}, \bibinfo{person}{Qihui Zhang}, \bibinfo{person}{Nuno Moniz}, \bibinfo{person}{Tian Gao}, \bibinfo{person}{Werner Geyer}, \bibinfo{person}{Chao Huang}, \bibinfo{person}{Pin-Yu Chen}, {et~al\mbox{.}}} \bibinfo{year}{2024}\natexlab{}.
\newblock \showarticletitle{Justice or prejudice? quantifying biases in llm-as-a-judge}.
\newblock \bibinfo{journal}{\emph{arXiv preprint arXiv:2410.02736}} (\bibinfo{year}{2024}).
\newblock


\bibitem[Yu et~al\mbox{.}(2023)]%
        {yu2023large}
\bibfield{author}{\bibinfo{person}{Yue Yu}, \bibinfo{person}{Yuchen Zhuang}, \bibinfo{person}{Jieyu Zhang}, \bibinfo{person}{Yu Meng}, \bibinfo{person}{Alexander~J Ratner}, \bibinfo{person}{Ranjay Krishna}, \bibinfo{person}{Jiaming Shen}, {and} \bibinfo{person}{Chao Zhang}.} \bibinfo{year}{2023}\natexlab{}.
\newblock \showarticletitle{Large language model as attributed training data generator: A tale of diversity and bias}.
\newblock \bibinfo{journal}{\emph{Advances in Neural Information Processing Systems}}  \bibinfo{volume}{36} (\bibinfo{year}{2023}), \bibinfo{pages}{55734--55784}.
\newblock


\bibitem[Zhang et~al\mbox{.}({[n.\,d.]})]%
        {zhangtask}
\bibfield{author}{\bibinfo{person}{Jieyu Zhang}, \bibinfo{person}{Weikai Huang}, \bibinfo{person}{Zixian Ma}, \bibinfo{person}{Oscar Michel}, \bibinfo{person}{Dong He}, \bibinfo{person}{Tanmay Gupta}, \bibinfo{person}{Wei-Chiu Ma}, \bibinfo{person}{Ali Farhadi}, \bibinfo{person}{Aniruddha Kembhavi}, {and} \bibinfo{person}{Ranjay Krishna}.} \bibinfo{year}{[n.\,d.]}\natexlab{}.
\newblock \showarticletitle{Task Me Anything}. In \bibinfo{booktitle}{\emph{The Thirty-eight Conference on Neural Information Processing Systems Datasets and Benchmarks Track}}.
\newblock


\bibitem[Zhang et~al\mbox{.}(2023)]%
        {zhang2023llmaaa}
\bibfield{author}{\bibinfo{person}{Ruoyu Zhang}, \bibinfo{person}{Yanzeng Li}, \bibinfo{person}{Yongliang Ma}, \bibinfo{person}{Ming Zhou}, {and} \bibinfo{person}{Lei Zou}.} \bibinfo{year}{2023}\natexlab{}.
\newblock \showarticletitle{LLMaAA: Making Large Language Models as Active Annotators}. In \bibinfo{booktitle}{\emph{Findings of the Association for Computational Linguistics: EMNLP 2023}}. \bibinfo{pages}{13088--13103}.
\newblock


\bibitem[Zhu et~al\mbox{.}(2023)]%
        {zhu2023dyval}
\bibfield{author}{\bibinfo{person}{Kaijie Zhu}, \bibinfo{person}{Jiaao Chen}, \bibinfo{person}{Jindong Wang}, \bibinfo{person}{Neil~Zhenqiang Gong}, \bibinfo{person}{Diyi Yang}, {and} \bibinfo{person}{Xing Xie}.} \bibinfo{year}{2023}\natexlab{}.
\newblock \showarticletitle{Dyval: Dynamic evaluation of large language models for reasoning tasks}.
\newblock \bibinfo{journal}{\emph{arXiv preprint arXiv:2309.17167}} (\bibinfo{year}{2023}).
\newblock


\bibitem[Zhu et~al\mbox{.}(2024)]%
        {zhu2024dyval}
\bibfield{author}{\bibinfo{person}{Kaijie Zhu}, \bibinfo{person}{Jindong Wang}, \bibinfo{person}{Qinlin Zhao}, \bibinfo{person}{Ruochen Xu}, {and} \bibinfo{person}{Xing Xie}.} \bibinfo{year}{2024}\natexlab{}.
\newblock \showarticletitle{Dyval 2: Dynamic evaluation of large language models by meta probing agents}.
\newblock \bibinfo{journal}{\emph{arXiv preprint arXiv:2402.14865}} (\bibinfo{year}{2024}).
\newblock


\end{thebibliography}

\end{document}